\documentclass[conference]{IEEEtran}
\usepackage{paper}
\IEEEoverridecommandlockouts

\def\BibTeX{{\rm B\kern-.05em{\sc i\kern-.025em b}\kern-.08em
    T\kern-.1667em\lower.7ex\hbox{E}\kern-.125emX}}
\begin{document}

\title{CSI-Based Efficient Self-Quarantine Monitoring System Using Branchy Convolution Neural Network}

\author{\IEEEauthorblockN{Jingtao Guo}
\IEEEauthorblockA{\textit{Dept. of Electronic and Information Engineering} \\
\textit{The Hong Kong Polytechnic University}\\
Hong Kong, China \\
\href{mailto:jingtao.guo@connect.polyu.hk}{\color{black}{jingtao.guo@connect.polyu.hk}}}
\and
\IEEEauthorblockN{Ivan Wang-Hei Ho}
\IEEEauthorblockA{\textit{Dept. of Electronic and Information Engineering} \\
\textit{Otto Poon Charitable Foundation Smart Cities Research Institute} \\
\textit{The Hong Kong Polytechnic University}\\
Hong Kong, China \\
\href{mailto:ivanwh.ho@polyu.edu.hk}{\color{black}{ivanwh.ho@polyu.edu.hk}}}
}


\maketitle

\begin{abstract}
Nowadays, Coronavirus disease (COVID-19) has become a global pandemic because of its fast spread in various countries. To build an anti-epidemic barrier, self-isolation is required for people who have been to any at-risk places or have been in close contact with infected people. However, existing camera or wearable device-based monitoring systems may present privacy leakage risks or cause user inconvenience in some cases. In this paper, we propose a Wi-Fi-based device-free self-quarantine monitoring system. Specifically, we exploit channel state information (CSI) derived from Wi-Fi signals as human activity features. We collect CSI data in a simulated self-quarantine scenario and present BranchyGhostNet, a lightweight convolution neural network (CNN) with an early exit prediction branch, for the efficient joint task of room occupancy detection (ROD) and human activity recognition (HAR). The early exiting branch is used for ROD, and the final one is used for HAR. Our experimental results indicate that the proposed model can achieve an average accuracy of 98.19\% for classifying five different human activities. They also confirm that after leveraging the early exit prediction mechanism, the inference latency for ROD can be significantly reduced by 54.04\% when compared with the final exiting branch while guaranteeing the accuracy of ROD.
\end{abstract}


\section{Introduction}
\label{sec: sec1}

In recent years, COVID-19 (an infectious disease caused by the SARS-CoV-2 virus) has become a global pandemic and caused a serious threat to our health \cite{whocovidpandemic}. Over one million cases have been reported in Hong Kong \cite{hkcovid19}, causing a heavy burden on the public health system. According to the existing findings \cite{whocovid19}, when infected people cough, sneeze, speak, or breath, the virus spreads in little liquid particles from their mouth or nose, and on average, it takes five to six days from the time people are infected with the virus to the time they develop symptoms, but it can take up to 14 days. Therefore, self-quarantine is required for those who have been in close contact with infected people or to any at-risk places to prevent the virus from spreading around the community. 

\subsection{Related Works and Motivations}

\textit{Strengths of Device-Free Wireless Sensing Based COVID-19 Monitoring System When Compared With Existing Systems:} Currently, several COVID-19 monitoring and tracing systems are developed by various countries and regions to control the spread of the COVID-19 virus for protecting the health of their community. For example, the Malaysian government has developed a contact tracing application termed MySejahtera\cite{MySejahtera} to trace users who have stayed with infected people on the same premises via a QR code check-in scheme. However, this application is unsuitable for monitoring someone who is undertaking self-quarantine since it serves as proactively preventative measures to alert users if they have been in close contact with infected people. Hong Kong government proposed a self-quarantine monitoring scheme named StayHomeSafe\cite{StayHomeSafe}. This scheme comprises an application and a wristband with a QR code for scanning at a random time to prevent quarantine users from leaving their place during the quarantine period. \cite{tan2021covid} also presented a self-quarantine monitoring system that used GPS signals and a camera to locate and verify a quarantine user, respectively. After that, \cite{Lim2022} introduced a monitoring system that can identify a quarantine user and periodically check the user's location and health status. However, camera-based monitoring systems usually require complex data pre-processing methods in dark scenarios. They also present privacy leakage risks. Monitoring systems with wearable devices may cause user inconvenience. Compared with the sensing technologies mentioned above, wireless sensing technology has great advantages in low-level privacy leakage, contactless operation, and well-functioning in low-light and non-line-of-sight (NLOS) environments\cite{Zhang2021}. 

\textit{Existing CSI-Based HAR and ROD Systems:} In the past decades, Wi-Fi CSI-based wireless sensing technology is gaining popularity as a result of the availability of different CSI tools like the Linux 802.11n CSI Tool\cite{halperin2011tool} and Nexmon CSI\cite{gringoli2019free}. In terms of these tools, several CSI-based wireless sensing systems have been proposed. For modeling the relationship between human activities and the dynamically changing signal features, machine learning (ML) technologies are applied in these systems. For example, \cite{Wang2017} used a one-class support vector machine (SVM) and random forest (RF) algorithms for HAR and ROD. \cite{Xin2016} also introduced a k-nearest neighbor (KNN) classifier for human identification. However, conventional ML approaches usually require hand-crafted feature extractors\cite{lecun1998gradient}. Hence, more researchers exploit deep learning (DL) algorithms for HAR and ROD.\begin{figure*}[htbp]
    \centering
    \includegraphics[width=1.55\columnwidth,height=0.31\linewidth]{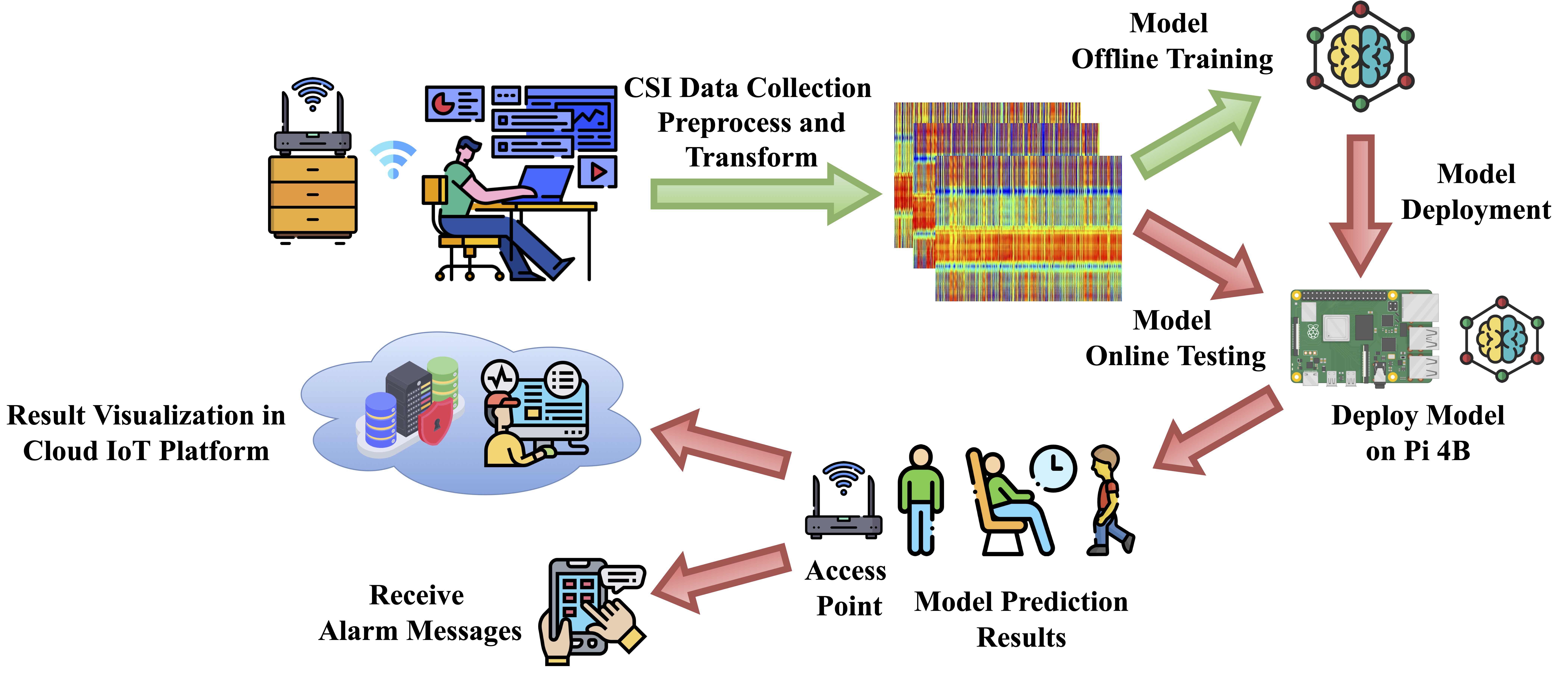}
    \caption{The overview of our proposed self-quarantine monitoring system.}
    \label{fig:pic1}
\end{figure*} DL technologies can regard as an end-to-end learning approach where feature extraction, a time-consuming and knowledge-demanding process in traditional ML algorithms, is accomplished automatically and embedded implicitly in the network architecture\cite{alhomayani2020deep}. For instance, \cite{tunet} proposed a temporal Unet (TUNet) for sample-level classification of different human activities. \cite{Wang2019} further leveraged the ResNet\cite{He2016} architecture to design a one-dimensional ResNet (1D-ResNet) for the joint task of HAR and indoor localization. \cite{Chen2019} also presented an attention-based bi-directional long short-term memory (ABLSTM) for passive HAR and showed satisfactory performance. After that, \cite{Forbes2020} combined the 1D-CNN and bi-directional LSTM (BiLSTM) architectures to present a deep convolutional LSTM (DeepConvLSTM) model and achieved 92\% average accuracy for eleven activities. \cite{Memmesheimer2020} exploited an efficient 2D-CNN named EfficientNet-B2\cite{efficientnet} for HAR using four action recognition datasets. \cite{Khan2022} also leveraged 2D-CNN and transfer learning to conduct ROD. In \cite{Zhuravchak2022}, InceptionTime\cite{IsmailFawaz2020} and BiLSTM-based models are introduced to evaluate their performance for HAR. However, these DL benchmarks either only consider single-task learning or do not achieve a good balance between accuracy and latency.

\subsection{Contributions}
In this paper, we propose a Wi-Fi-based self-quarantine monitoring system. Specifically, we utilize CSI extracted from Wi-Fi signals as human activity features. Since ROD is an easy task than HAR in a self-quarantine scenario, we introduce the early exit prediction mechanism to GhostNet\cite{han2020ghostnet} model to design a new DL model for the joint task of HAR and ROD. This mechanism can significantly reduce the inference time of the model for ROD without compromising the detection accuracy. The prediction result is sent to a cloud IoT platform for visualization and remote monitoring. Compared with other monitoring systems mentioned above, our proposed method has great advantages in low-level privacy leakage and contact-free monitoring. Real-world experiments have been conducted to verify the efficiency of the proposed model. The experimental results are compared with the other seven DL benchmarks presented in the literature. Overall, the main contributions of this study are three-fold:
\begin{itemize}
\item We present a Wi-Fi-based self-quarantine monitoring system to detect human activities and room occupancy where CSI data extracted from Wi-Fi signals are exploited as human activity features. We also introduce a demonstration leveraging the Raspberry Pi 4B, ThingsBoard, and Telegram platforms.
\item To improve the efficiency for ROD, we then introduce BranchyGhostNet for the joint task of HAR and ROD in which the early exit prediction mechanism can significantly reduce the inference time of the model for ROD without compromising the detection accuracy.
\item We conduct extensive experiments to evaluate our model. Our results indicate that the proposed BranchyGhostNet can achieve up to 13.22\% performance boost for HAR compared with the other six common-used DL models and reduce the inference time for ROD by 54.04\% compared with its final exiting branch.
\end{itemize}

The remaining part of the paper proceeds as follows: In Section \ref{sec: sec2}, we illustrate our proposed monitoring system and BranchyGhostNet. After that, we introduce the experimental setup for evaluating the proposed system in Section \ref{sec: sec3}. Section \ref{sec: sec4} presents the experimental results. Finally, Section \ref{sec: sec5} concludes the paper.

\section{The Proposed Self-Quarantine Monitoring System and BranchyGhostNet}
\label{sec: sec2}
\subsection{Main Idea of the proposed system}

Our proposed monitoring system aims to detect quarantine user activities and prevent the user from leaving the quarantine place or being in close contact with another person. Fig.~\ref{fig:pic1} demonstrates the overview of our system. We suppose a user is undertaking self-quarantine. The following three parts comprise our monitoring system. $\left(1\right)$ Train a lightweight model using the CSI data collected in the simulated self-quarantine scenario. $\left(2\right)$ The trained model is converted and deployed to the Raspberry Pi 4B for online testing. $\left(3\right)$ In the online testing stage, the prediction result is sent to a cloud IoT platform for remote monitoring, and an alarm message is sent to the messenger if the room is empty or the user comes into contact with another person.
\begin{figure*}[htbp]
    \centering
    \includegraphics[width=2.04\columnwidth,height=0.37\linewidth]{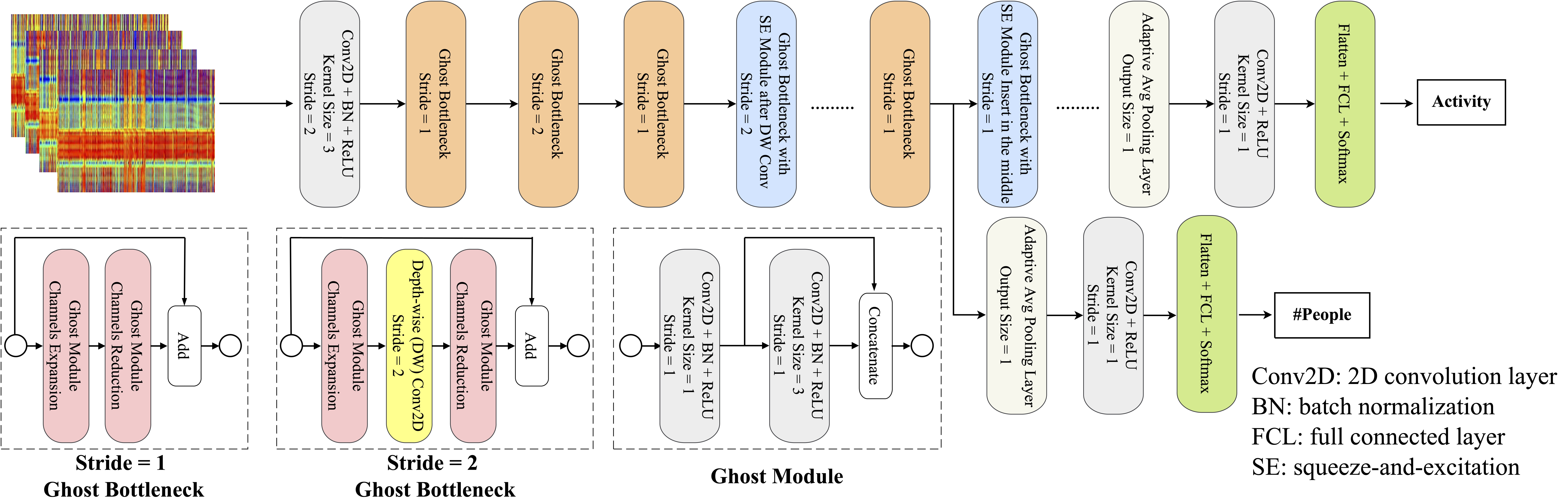}
    \caption{The architecture of the proposed BranchyGhostNet.}
    \label{fig:pic2}
\end{figure*}

The main computational procedure in our proposed system comes from offline model training. We exploit a lightweight model named GhostNet\cite{han2020ghostnet} to design a new DL model named BranchyGhostNet for the joint task of HAR and ROD. This model is quite suitable for edge devices because of its low computational cost and inference latency. The learning goal for user $\boldsymbol{p}$'s model is represented as

\begin{equation}
    \label{eqn_2}
	\mathop{\arg\min_{\Theta^{\boldsymbol{p}}}} \mathcal{L}_{\boldsymbol{p}} = \sum_{i=1}^{n^{\boldsymbol{p}}}\mathcal{C}\left(\boldsymbol{l}_{i\_r}^{\boldsymbol{p}},\boldsymbol{B}_{e}^{\boldsymbol{p}}\left(\boldsymbol{X}_{i}^{\boldsymbol{p}}\right)\right) + \sum_{i=1}^{n^{\boldsymbol{p}}}\mathcal{C}\left(\boldsymbol{l}_{i\_h}^{\boldsymbol{p}},\boldsymbol{B}_{o}^{\boldsymbol{p}}\left(\boldsymbol{X}_{i}^{\boldsymbol{p}}\right)\right)
\end{equation}

where $\Theta^{\boldsymbol{p}}$ means the optimal parameters of all layers for user $\boldsymbol{p}$’s model, $\mathcal{C}\left(\cdot,\cdot\right)$ denotes as the loss function for model training (e.g., cross-entropy loss function), $\mathcal{L}_{\boldsymbol{p}}$ means the overall loss for model training. $\boldsymbol{B}_{e}^{\boldsymbol{p}}$ and $\boldsymbol{B}_{o}^{\boldsymbol{p}}$ represent the early exit and final exit prediction branch of $\boldsymbol{p}$’s model, respectively. $\left\{\boldsymbol{X}_{i}^{\boldsymbol{p}},\ \boldsymbol{l}_{i\_r}^{\boldsymbol{p}},\ \boldsymbol{l}_{i\_h}^{\boldsymbol{p}}\right\}_{i=1}^{n^{\boldsymbol{p}}}$ are samples and their corresponding true ROD and HAR labels from $\boldsymbol{p}$’s dataset with size $\boldsymbol{n^{\boldsymbol{p}}}$. 

\subsection{Model Architecture}
A deeper structure is usually required for a DL model to achieve higher accuracy. However, the cost of added latency and energy usage becomes more prohibitive for real-time and energy-sensitive applications when the model continues to get deeper and larger. Since ROD is an easy task than HAR under the self-quarantine scenario and inspired by \cite{Teerapittayanon2016}, we introduce the early exit prediction mechanism to the original GhostNet model to conduct the joint task of HAR and ROD while reducing the inference latency for ROD to improve the monitoring efficiency. This mechanism utilizes the early layer of a DL model to learn the hidden features for ROD, which indicates that the early exit prediction branch shares part of the layers with the final prediction branch and allows certain prediction results to exit from the model early. With this mechanism, our proposed BranchyGhostNet structure is presented by adding an exit branch after a certain layer position of the original GhostNet model as shown in Fig.~\ref{fig:pic2}. The BranchyGhostNet model comprises 97 convolution layers and two fully connected layers with only about 370 million multiply-accumulate operations (MACs). The main component of our model is the Ghost module, which mainly uses cheap linear operation to augment features and generate more feature maps after obtaining a few inherent channels by a regular convolution operation\cite{han2020ghostnet}. The Ghost bottleneck with stride=1 comprises two Ghost modules, and that with stride=2 further includes a depthwise convolution layer with stride=2 between the two Ghost modules. The squeeze-and-excitation module, which can utilize the correlation between feature maps, is also adopted in some Ghost bottlenecks.

\begin{figure}[htbp]
    \centering
    \includegraphics[scale=0.04]{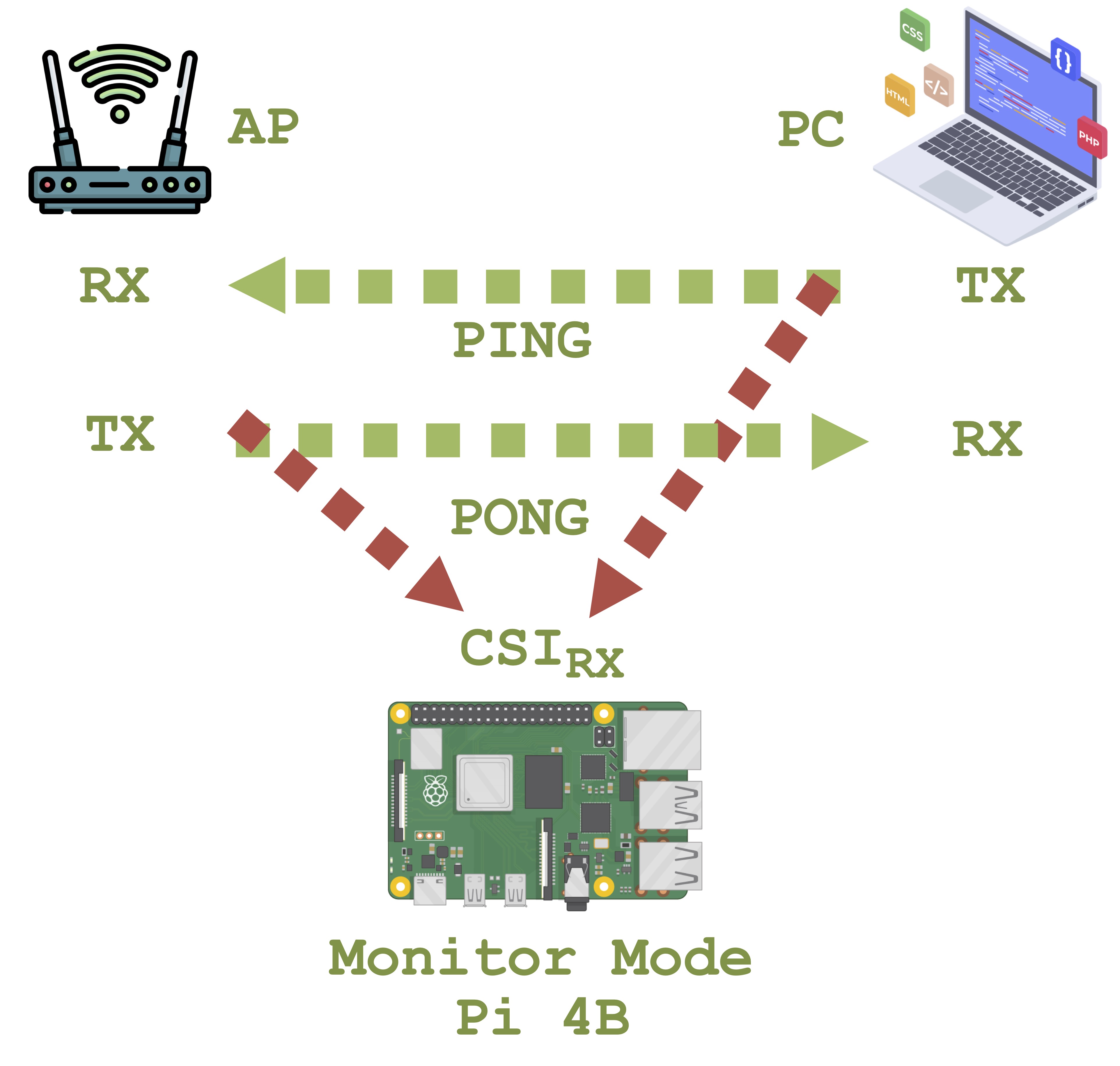}
    \caption{CSI data collection method.}
    \label{fig:pic3}
\end{figure}

\section{Experiment Setup}
\label{sec: sec3}
\subsection{Equipment and Data Collection Method}
Fig.~\ref{fig:pic3} illustrates the communication between the devices for CSI data collection. The PC is used to send control messages to the AP for generating the wireless signals that contain CSI data, while a Raspberry Pi 4B serves as a data collector to extract CSI data. Ping packets are sent from the PC to the AP, and pong packets are sent back from the AP to the PC. 

The Pi 4B in monitor mode was configured with Raspberry Pi OS (Buster/Linux 5.4.83) with the pi-5.4.51-plus branch of nexmon\_csi\footnote{https://github.com/zeroby0/nexmon\_csi/tree/pi-5.4.51-plus} installed. The following filter parameters were used for configuring nexmon: channel 36/80, Core 1, NSS mask 1. The Raspberry Pi 4B was paired to the AP. A computer was connected to the Pi 4B over the SSH protocol on a separate 2.4 GHz channel to control the data collection and reduce interference. The AP operates on channel 36 with an 80 MHz bandwidth. Note that the model of the AP is not restricted as long as it supports the 802.11ac protocol. Finally, the PC is connected to the 5 GHz channel of the AP to generate data flow from which the Pi 4B in monitor mode can capture CSI data. The PC is configured to send ping flood to the AP at a rate of 100 Hz.

\begin{figure}[htbp]
    \centering
    \includegraphics[scale=0.078]{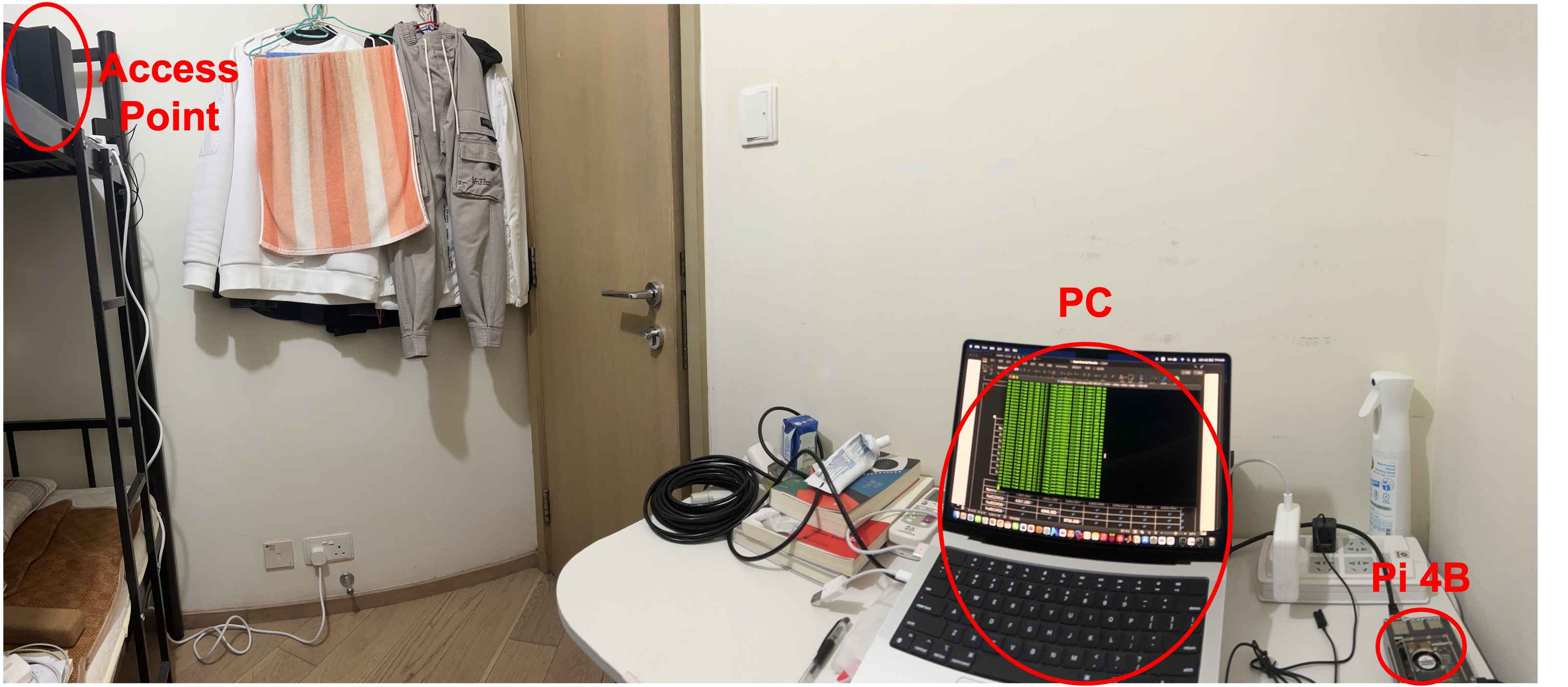}
    \caption{The layout of simulated self-quarantine environment.}
    \label{fig:pic4}
\end{figure}

\subsection{Environments and Datasets}

Fig.~\ref{fig:pic4} shows the layout of an environment which was considered for CSI data collection. In this environment, we used the nexmon CSI patch to collect five human activities (sit, stand, walk, stand up and sit down) and three scenarios of room occupancy (nobody, one person, and two persons). The tcpdump command is used to render CSI data for producing a pcap file. An open-source framework named csiread is used to interpret this file, which generates a 256 x 300 CSI amplitude matrix after computing the absolute values and the transpose operation. Note that 256 means the number of sub-carriers, and 300 represents the number of received data packets. Overall, datasets contain 10,000 CSI data for training and 2,300 CSI data for testing.

\subsection{Data Preprocessing}
Since signal preprocessing can degrade the timeliness of a real-time system, only the amplitude response of CSI is utilized in this study. We minimize the signal preprocessing in our system. First, 8 Pilot sub-carriers and 14 Null sub-carriers (including Guard sub-carriers) are filtered away according to \cite{gast2013802}. Then, we adopt the moving average method to eliminate short-term fluctuations of the signals and highlight their long-term trends. Finally, we save them as CSI radio images with 234 x 300 resolution to utilize the spatial and temporal correlations between adjacent channels and samples\cite{wang2018device}.

\subsection{Model Training and Testing Configuration}

Our model is implemented by Pytorch and trained using a computer configured with Nvidia RTX 3090 GPU. Before training the model, we use some general data argumentation methods (RandomRiszedCrop, RandomHorizontalFlip, and ColorJitter) when loading our training data. The AdamW optimizer is used to optimize model parameters during the training phase. CosineAnnealingLR scheduler is also applied to attenuate the learning rate via the cosine function, and a cross-entropy loss function is adopted for computing the training loss. The training epoch, training batch size, and testing batch size are set to 400, 50, and one, respectively. For the early exit prediction criterion, we set that the prediction result will output early from the model if it is equal to zero or two. In our experiments, four criteria, i.e., accuracy, precision, recall, and f1-score\cite{Rocamora2020}, are used to evaluate the performance of the BranchyGhostNet. We also assess its inference latency in GPU and compare it with the other seven DL benchmarks.

\section{Experiment Evaluation}
\label{sec: sec4}
In this section, we use CSI data collected by the Raspberry Pi 4B platform via Wi-Fi channels from a single router as human activity features to validate the performance of our model and compare it with seven other common-used DL benchmarks. We also utilize this model to build a Wi-Fi-based self-quarantine monitoring system. 

\subsection{The performance of BranchyGhostNet}

Table~\ref{tab:tab1} and \ref{tab:tab2} show the accuracy, precision, recall, f1-score and mean GPU latency of eight different DL models. As can be seen from the tables below, EfficientNet-B2 presents the best performance among eight DL models for both ROD and HAR. However, it provides the largest latency, which may not be suitable for resource-constrained devices and latency-sensitive tasks. Compared with EfficientNet-B2, our model only has a 0.58\% and 0.23\% accuracy drop with 73.82\% and 43.03\% latency reduction for ROD and HAR, respectively. In comparison with the mean latency of BranchyGhostNet shown in Table~\ref{tab:tab2}, that presented in Table~\ref{tab:tab1} indicates that after introducing the early exit prediction mechanism for ROD, an about two times computation speed-up ratio can be achieved without comprising the accuracy. From the results in Table~\ref{tab:tab1} and \ref{tab:tab2}, we can also observe that most of the LSTM-based models have worse performance than CNN-based models, which implies a strong feature extraction capability of CNN for CSI-based HAR and ROD. \begin{figure}[htbp]
    \centering
    \includegraphics[scale=0.35]{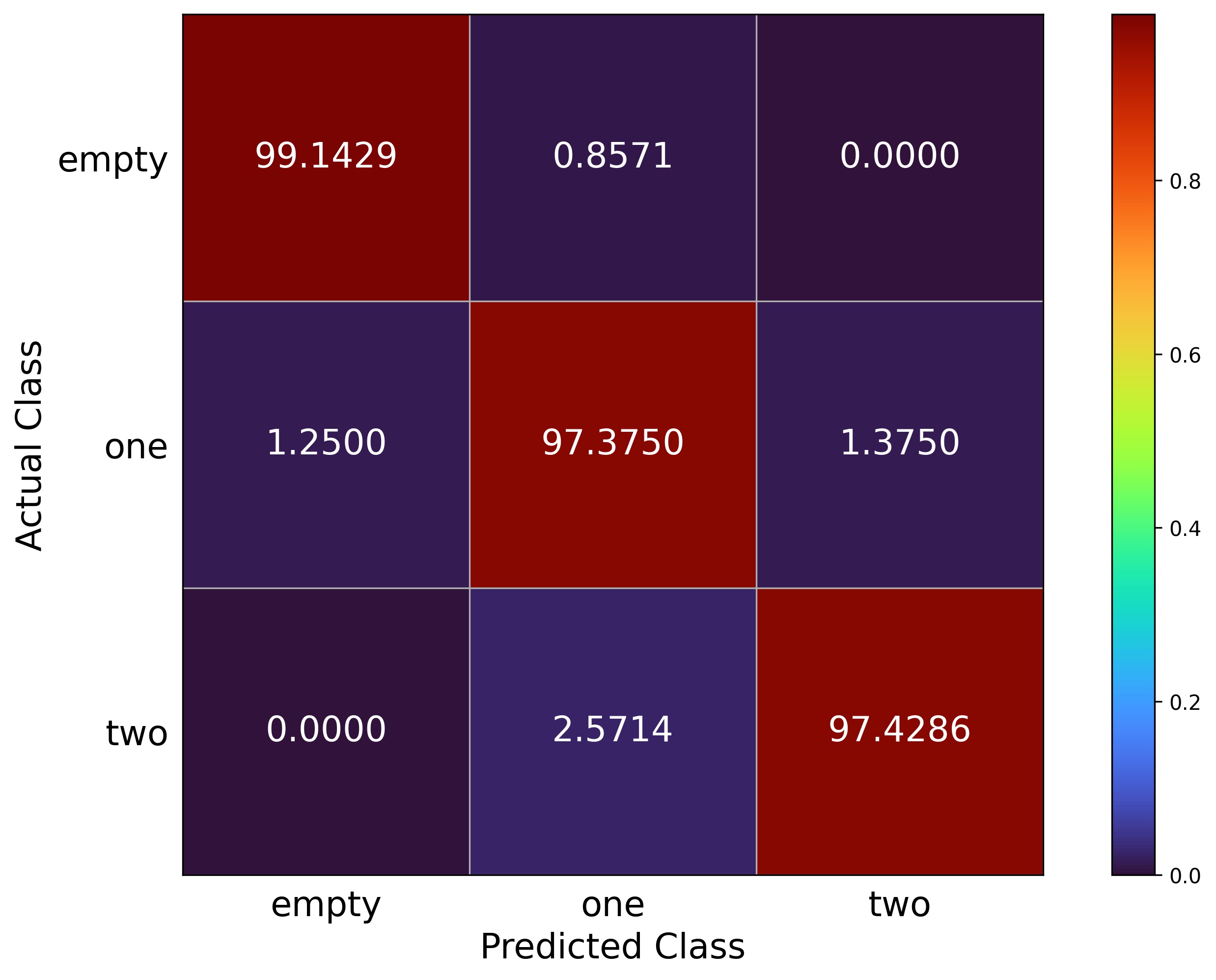}
    \caption{Confusion matrix (\%) of BranchyGhostNet for ROD.}
    \label{fig:pic5}
\end{figure}
\begin{table*}[htbp]
\centering
\caption{Accuracy, precision, recall, f1-score (\%) and mean GPU inference latency (ms) for ROD}\label{tab:tab1}
\begin{tabular}{cccccc}
\toprule
\textbf{Model} & \textbf{Accuracy} & \textbf{Precision} & \textbf{Recall} & \textbf{F1-score} & \textbf{Mean inference latency} \\\midrule 
BranchyGhostNet & \textbf{98.43} & \textbf{97.65} & \textbf{97.65} & \textbf{97.65} & \textbf{8.13} \\
EfficientNet-B2\cite{Memmesheimer2020} & 99.01 & 98.52 & 98.52 & 98.52 & 31.05 \\
ABLSTM\cite{Chen2019} & 95.86 & 93.78 & 93.78 & 93.78 & 10.53 \\
BiLSTM-based Model\cite{Zhuravchak2022} & 97.51 & 96.26 & 96.26 & 96.26 & 10.44 \\
1D-ResNet\cite{Wang2019} & 96.49 & 94.74 & 94.74 & 94.74 & 6.17 \\
DeepConvLSTM\cite{Forbes2020} & 96.32 & 94.48 & 94.48 & 94.48 & 20.53 \\
InceptionTime\cite{IsmailFawaz2020} & 98.20 & 97.30 & 97.30 & 97.30 & 23.08 \\
TUNet\cite{tunet} & 96.87 & 95.30 & 95.30 & 95.30 & 9.53 \\
\bottomrule
\end{tabular}
\end{table*}
\begin{table*}[htbp]
\centering
\caption{Accuracy, precision, recall, f1-score (\%) and mean GPU inference latency (ms) for HAR}\label{tab:tab2}
\begin{tabular}{cccccc}
\toprule
\textbf{Model} & \textbf{Accuracy} & \textbf{Precision} & \textbf{Recall} & \textbf{F1-score} & \textbf{Mean inference latency} \\\midrule 
BranchyGhostNet & \textbf{98.19} & \textbf{95.48} & \textbf{95.48} & \textbf{95.48} & \textbf{17.69} \\
EfficientNet-B2\cite{Memmesheimer2020} & 98.42 & 96.04 & 96.04 & 96.04 & 31.05 \\
ABLSTM\cite{Chen2019} & 92.90 & 82.26 & 82.26 & 82.26 & 10.53 \\
BiLSTM-based Model\cite{Zhuravchak2022} & 95.22 & 88.04 & 88.04 & 88.04 & 10.44 \\
1D-ResNet\cite{Wang2019} & 96.45 & 91.13 & 91.13 & 91.13 & 6.17 \\
DeepConvLSTM\cite{Forbes2020} & 95.70 & 89.26 & 89.26 & 89.26 & 20.53 \\
InceptionTime\cite{IsmailFawaz2020} & 95.76 & 89.39 & 89.39 & 89.39 & 23.08 \\
TUNet\cite{tunet} & 96.31 & 90.78 & 90.78 & 90.78 & 9.53 \\
\bottomrule
\end{tabular}
\end{table*}

The confusion matrices illustrating ROD and HAR accuracy of our proposed model are presented in Fig.~\ref{fig:pic5} and \ref{fig:pic6} respectively. From the two confusion matrices shown below, we can see that higher accuracies are obtained for ROD. That may be because the number of classes in HAR is more than that in ROD, resulting in more powerful shared layers for the model and strong feature extraction capability for ROD. It can be seen from the confusion matrix shown in Figure~\ref{fig:pic6} that all activities except the "walk" activity work well, while the "walk" activity shows a slight drop with lower than 90\% accuracy. This activity will cause more fluctuation for the received signals, which may produce more noisy data patterns for the model.
\begin{figure}[htbp]
    \centering
    \includegraphics[scale=0.35]{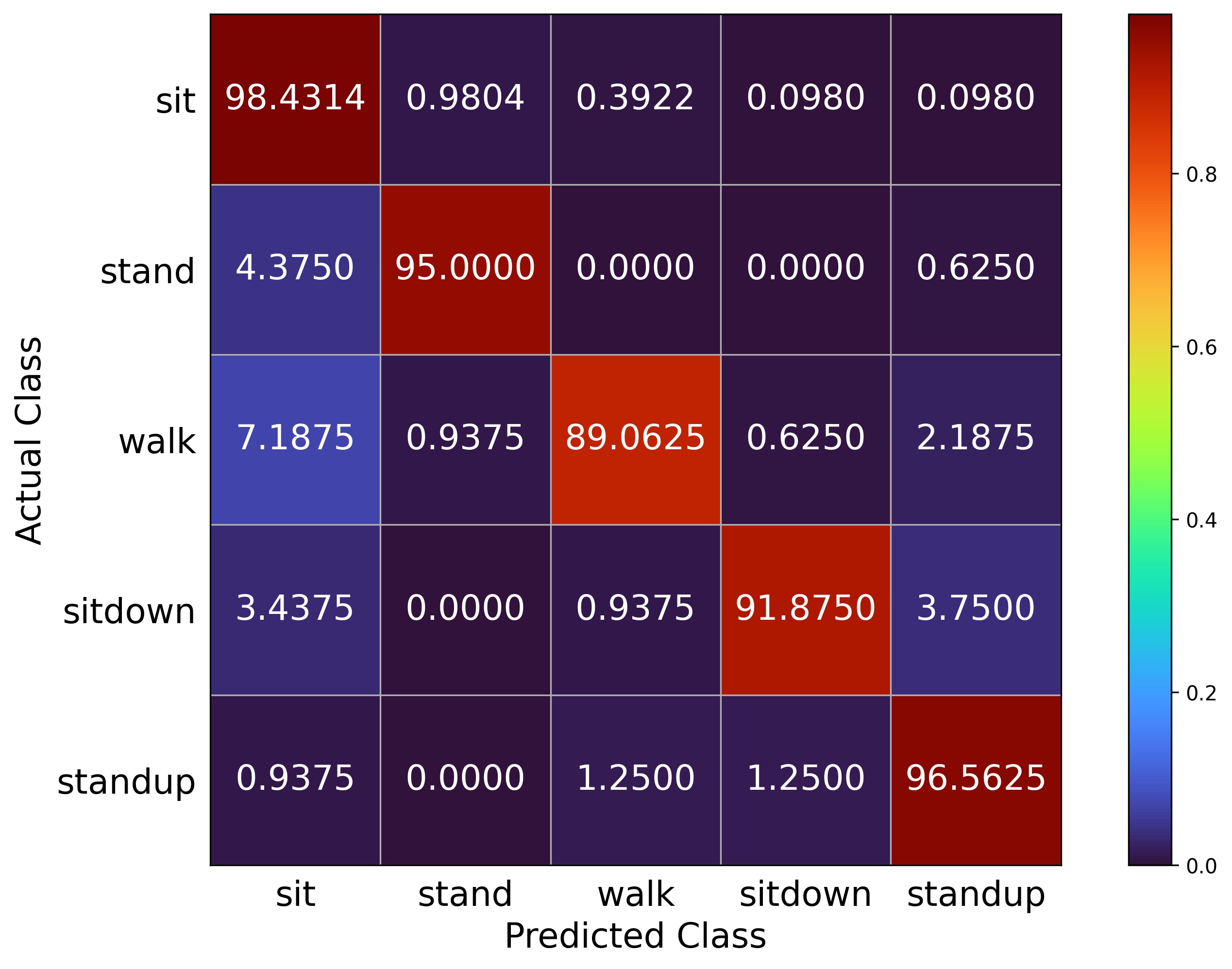}
    \caption{Confusion matrix (\%) of BranchyGhostNet for HAR.}
    \label{fig:pic6}
\end{figure} 
\subsection{Demonstration of the proposed self-quarantine monitoring system}

\begin{figure}[htbp]
    \centering
    \includegraphics[scale=0.167]{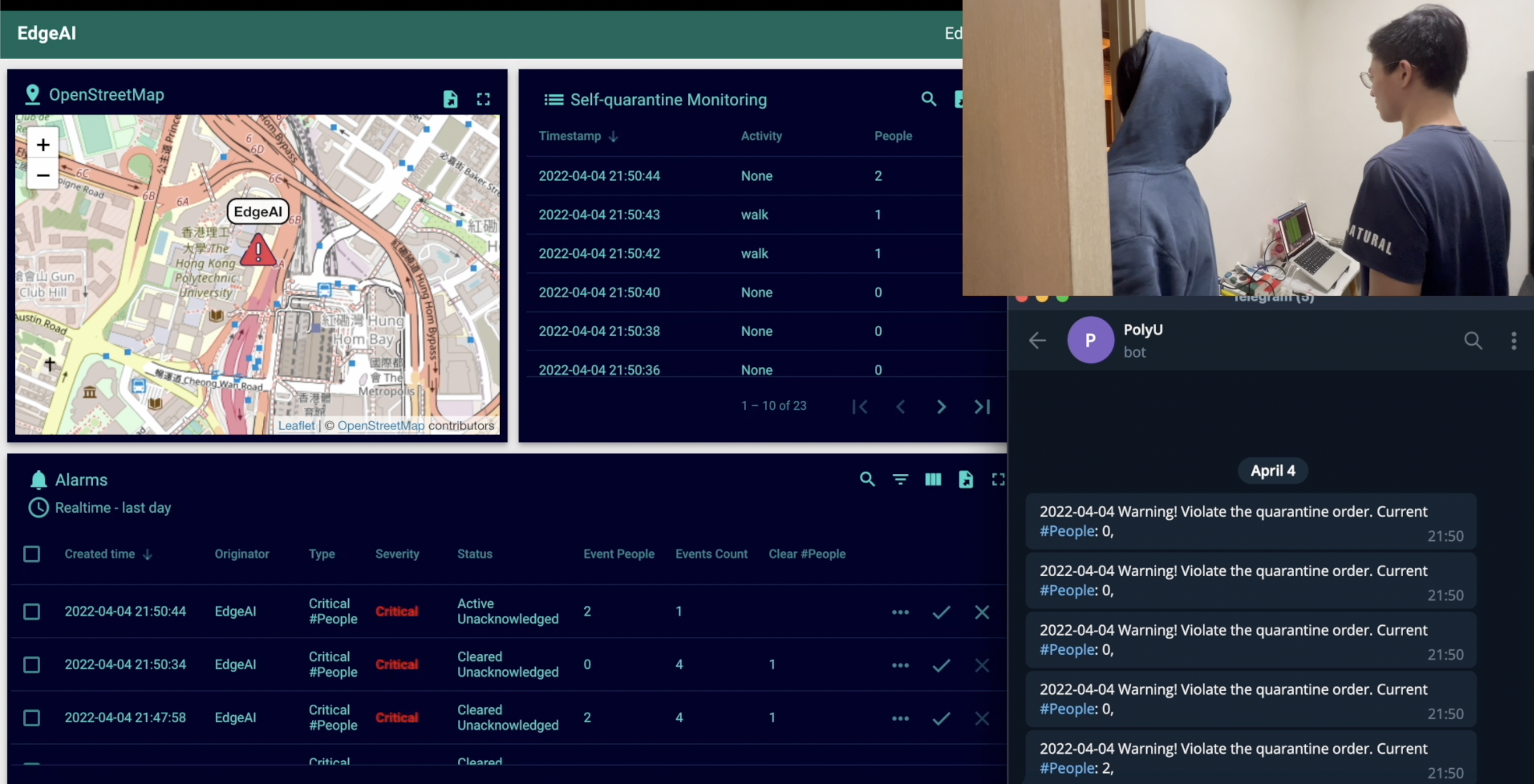}
    \caption{Detection result visualization and alarm message receiving in the cloud IoT platform and messenger respectively.}
    \label{fig:pic7}
\end{figure}

By exploiting the proposed BranchyGhostNet, we present a Wi-Fi CSI-based self-quarantine monitoring system. Fig.~\ref{fig:pic7} shows the detection result visualization and alarm message receiving in ThingsBoard (cloud internet of things (IoT) platform) and Telegram (messenger) respectively to remotely monitor. Since the Raspberry Pi 4B only has a CPU, we further converted the model from the Pytorch framework into the open neural network exchange (ONNX) framework to reduce the inference latency in the CPU.
 
\section{Conclusion and Future Work}
\label{sec: sec5}
This paper proposed a CSI-based device-free system for monitoring a user during the quarantine period while preserving the user's privacy. To improve the efficiency of ROD, we introduced the early exit prediction mechanism to an original GhostNet to design a BranchyGhostNet model for the joint task of HAR and ROD. We have also developed a graphical user interface (GUI) to visualize the prediction result and used a messenger to receive the warnings. We have evaluated the performance of our proposed model. The experimental results showed that our model can achieve the best balance between latency and accuracy (i.e., provide low latency while guaranteeing high accuracy) compared with the other seven common-used DL benchmarks. After introducing the early exit prediction mechanism for ROD in BranchyGhostNet, its latency can be reduced by 54.04\% when compared with the final exiting branch. 

A limitation of this study is that we currently only collect predefined activity data under an ideal scenario to train and test the model. For future research, how to handle unseen activities without the extra effort of data collection and model retraining can be investigated. We will also explore how to leverage unlabeled CSI data for model training to reduce the need for data labeling. 

\section{Acknowledgment}

This work was supported in part by the Key-Area Research and Development Program of Guangdong Province (2020B 090928001); and by The Hong Kong Polytechnic University (Project No. 4-ZZMU, Q-CDAS).



\begin{thebibliography}{10}
\providecommand{\url}[1]{#1}
\csname url@samestyle\endcsname
\providecommand{\newblock}{\relax}
\providecommand{\bibinfo}[2]{#2}
\providecommand{\BIBentrySTDinterwordspacing}{\spaceskip=0pt\relax}
\providecommand{\BIBentryALTinterwordstretchfactor}{4}
\providecommand{\BIBentryALTinterwordspacing}{\spaceskip=\fontdimen2\font plus
\BIBentryALTinterwordstretchfactor\fontdimen3\font minus
  \fontdimen4\font\relax}
\providecommand{\BIBforeignlanguage}[2]{{%
\expandafter\ifx\csname l@#1\endcsname\relax
\typeout{** WARNING: IEEEtran.bst: No hyphenation pattern has been}%
\typeout{** loaded for the language `#1'. Using the pattern for}%
\typeout{** the default language instead.}%
\else
\language=\csname l@#1\endcsname
\fi
#2}}
\providecommand{\BIBdecl}{\relax}
\BIBdecl

\bibitem{whocovidpandemic}
{World Health Organization}, ``Coronavirus disease (covid-19) weekly
  epidemiological update and weekly operational update,'' 2020, [Online].
  Available:
  \url{https://www.who.int/emergencies/diseases/novel-coronavirus-2019/situation-reports}
  [Accessed 31 May 2022].

\bibitem{hkcovid19}
{The Government of the Hong Kong Special Administrative Region}, ``Together,
  {We} {Fight} the {Virus},'' 2020, [Online]. Available:
  \url{https://www.coronavirus.gov.hk/sim/index.html} [Accessed 31 May 2022].

\bibitem{whocovid19}
{World Health Organization}, ``Coronavirus disease {(COVID-19)},'' 2020,
  [Online]. Available:
  \url{https://www.who.int/health-topics/coronavirus#tab=tab_1} [Accessed 31
  May 2022].

\bibitem{MySejahtera}
{The Government of Malaysia}, ``Mysejahtera,'' 2020, [Online]. Available:
  \url{https://mysejahtera.malaysia.gov.my/intro/} [Accessed 31 May 2022].

\bibitem{StayHomeSafe}
{The Government of the Hong Kong Special Administrative Region},
  ``{“StayHomeSafe”} {Mobile} {App} {User} {Guide},'' 2020, [Online].
  Available: \url{https://www.coronavirus.gov.hk/eng/stay-home-safe.html}
  [Accessed 31 May 2022].

\bibitem{tan2021covid}
Z.~F. Tan and W.~N.~W. Zakaria, ``{COVID-19} mandatory self-quarantine
  monitoring system,'' \emph{Evolution in Electrical and Electronic
  Engineering}, vol.~2, no.~2, pp. 605--614, 2021.

\bibitem{Lim2022}
W.~J. Lim and N.~M.~A. Ghani, ``{COVID}-19 mandatory self-quarantine wearable
  device for authority monitoring with edge {AI} reporting {\&} flagging
  system,'' \emph{Health and Technology}, vol.~12, no.~1, pp. 215--226, Jan.
  2022.

\bibitem{Zhang2021}
R.~Zhang, X.~Jing, S.~Wu, C.~Jiang, J.~Mu, and F.~R. Yu, ``{Device-Free}
  wireless sensing for human detection: The deep learning perspective,''
  \emph{{IEEE} Internet of Things Journal}, vol.~8, no.~4, pp. 2517--2539, Feb.
  2021.

\bibitem{halperin2011tool}
D.~Halperin, W.~Hu, A.~Sheth, and D.~Wetherall, ``Tool release: gathering
  802.11n traces with channel state information,'' \emph{{ACM} {SIGCOMM}
  Computer Communication Review}, vol.~41, no.~1, pp. 53--53, Jan. 2011.

\bibitem{gringoli2019free}
F.~Gringoli, M.~Schulz, J.~Link, and M.~Hollick, ``Free your {CSI}: A channel
  state information extraction platform for modern wi-fi chipsets,'' in
  \emph{Proceedings of the 13th International Workshop on Wireless Network
  Testbeds, Experimental Evaluation {\&} Characterization - {WiNTECH}
  {\textquotesingle}19}.\hskip 1em plus 0.5em minus 0.4em\relax {ACM} Press,
  2019.

\bibitem{Wang2017}
Y.~Wang, K.~Wu, and L.~M. Ni, ``{WiFall}: Device-free fall detection by
  wireless networks,'' \emph{{IEEE} Transactions on Mobile Computing}, vol.~16,
  no.~2, pp. 581--594, Feb. 2017.

\bibitem{Xin2016}
T.~Xin, B.~Guo, Z.~Wang, M.~Li, Z.~Yu, and X.~Zhou, ``{FreeSense}: Indoor human
  identification with wi-fi signals,'' in \emph{2016 {IEEE} Global
  Communications Conference ({GLOBECOM})}.\hskip 1em plus 0.5em minus
  0.4em\relax {IEEE}, Dec. 2016.

\bibitem{lecun1998gradient}
Y.~Lecun, L.~Bottou, Y.~Bengio, and P.~Haffner, ``Gradient-based learning
  applied to document recognition,'' \emph{Proceedings of the {IEEE}}, vol.~86,
  no.~11, pp. 2278--2324, 1998.

\bibitem{alhomayani2020deep}
F.~Alhomayani and M.~H. Mahoor, ``Deep learning methods for fingerprint-based
  indoor positioning: a review,'' \emph{Journal of Location Based Services},
  vol.~14, no.~3, pp. 129--200, Jul. 2020.

\bibitem{tunet}
\BIBentryALTinterwordspacing
F.~Wang, Y.~Song, J.~Zhang, J.~Han, and D.~Huang, ``Temporal unet: Sample level
  human action recognition using wifi,'' 2019. [Online]. Available:
  \url{https://arxiv.org/abs/1904.11953}
\BIBentrySTDinterwordspacing

\bibitem{Wang2019}
F.~Wang, J.~Feng, Y.~Zhao, X.~Zhang, S.~Zhang, and J.~Han, ``Joint activity
  recognition and indoor localization with {WiFi} fingerprints,'' \emph{{IEEE}
  Access}, vol.~7, pp. 80\,058--80\,068, 2019.

\bibitem{He2016}
K.~He, X.~Zhang, S.~Ren, and J.~Sun, ``Deep residual learning for image
  recognition,'' in \emph{2016 {IEEE} Conference on Computer Vision and Pattern
  Recognition ({CVPR})}.\hskip 1em plus 0.5em minus 0.4em\relax {IEEE}, Jun.
  2016.

\bibitem{Chen2019}
Z.~Chen, L.~Zhang, C.~Jiang, Z.~Cao, and W.~Cui, ``{WiFi} {CSI} based passive
  human activity recognition using attention based {BLSTM},'' \emph{{IEEE}
  Transactions on Mobile Computing}, vol.~18, no.~11, pp. 2714--2724, Nov.
  2019.

\bibitem{Forbes2020}
G.~Forbes, S.~Massie, and S.~Craw, ``{WiFi}-based human activity recognition
  using raspberry pi,'' in \emph{2020 {IEEE} 32nd International Conference on
  Tools with Artificial Intelligence ({ICTAI})}.\hskip 1em plus 0.5em minus
  0.4em\relax {IEEE}, Nov. 2020.

\bibitem{Memmesheimer2020}
R.~Memmesheimer, N.~Theisen, and D.~Paulus, ``Gimme signals: Discriminative
  signal encoding for multimodal activity recognition,'' in \emph{2020
  {IEEE}/{RSJ} International Conference on Intelligent Robots and Systems
  ({IROS})}.\hskip 1em plus 0.5em minus 0.4em\relax {IEEE}, Oct. 2020.

\bibitem{efficientnet}
\BIBentryALTinterwordspacing
M.~Tan and Q.~V. Le, ``Efficientnet: Rethinking model scaling for convolutional
  neural networks,'' 2019. [Online]. Available:
  \url{https://arxiv.org/abs/1905.11946}
\BIBentrySTDinterwordspacing

\bibitem{Khan2022}
D.~Khan and I.~W.-H. Ho, ``{CrossCount}: Efficient device-free crowd counting
  by leveraging transfer learning,'' \emph{{IEEE} Internet of Things Journal},
  pp. 1--1, 2022.

\bibitem{Zhuravchak2022}
A.~Zhuravchak, O.~Kapshii, and E.~Pournaras, ``Human activity recognition based
  on wi-fi {CSI} data - a deep neural network approach,'' \emph{Procedia
  Computer Science}, vol. 198, pp. 59--66, 2022.

\bibitem{IsmailFawaz2020}
H.~I. Fawaz, B.~Lucas, G.~Forestier, C.~Pelletier, D.~F. Schmidt, J.~Weber,
  G.~I. Webb, L.~Idoumghar, P.-A. Muller, and F.~Petitjean, ``{InceptionTime}:
  Finding {AlexNet} for time series classification,'' \emph{Data Mining and
  Knowledge Discovery}, vol.~34, no.~6, pp. 1936--1962, Sep. 2020.

\bibitem{han2020ghostnet}
K.~Han, Y.~Wang, Q.~Tian, J.~Guo, C.~Xu, and C.~Xu, ``{GhostNet}: More features
  from cheap operations,'' in \emph{2020 {IEEE}/{CVF} Conference on Computer
  Vision and Pattern Recognition ({CVPR})}.\hskip 1em plus 0.5em minus
  0.4em\relax {IEEE}, Jun. 2020.

\bibitem{Teerapittayanon2016}
S.~Teerapittayanon, B.~McDanel, and H.~Kung, ``{BranchyNet}: Fast inference via
  early exiting from deep neural networks,'' in \emph{2016 23rd International
  Conference on Pattern Recognition ({ICPR})}.\hskip 1em plus 0.5em minus
  0.4em\relax {IEEE}, Dec. 2016.

\bibitem{gast2013802}
M.~S. Gast, \emph{802.11 ac: a survival guide: Wi-Fi at gigabit and
  beyond}.\hskip 1em plus 0.5em minus 0.4em\relax " O'Reilly Media, Inc.",
  2013.

\bibitem{wang2018device}
J.~Wang, L.~Zhang, Q.~Gao, M.~Pan, and H.~Wang, ``Device-free wireless sensing
  in complex scenarios using spatial structural information,'' \emph{IEEE
  Transactions on Wireless Communications}, vol.~17, no.~4, pp. 2432--2442,
  2018.

\bibitem{Rocamora2020}
J.~M. Rocamora, I.~W.-H. Ho, W.-M. Mak, and A.~P.-T. Lau, ``Survey of {CSI}
  fingerprinting-based indoor positioning and mobility tracking systems,''
  \emph{{IET} Signal Processing}, vol.~14, no.~7, pp. 407--419, Sep. 2020.

\end{thebibliography}

\end{document}